\documentclass[10pt,journal,compsoc]{IEEEtran}
\usepackage[nocompress]{cite}
\usepackage[pdftex]{graphicx}
\graphicspath{{figures/}{author/}}
\DeclareGraphicsExtensions{.pdf,.jpeg,.png}
\usepackage{amsmath,bm}
\usepackage{algorithmic}
\usepackage{array}
\usepackage{url}
\usepackage[x11names]{xcolor}
\usepackage{tabulary,xfrac,enumitem}
\usepackage{booktabs}       
\usepackage{amsfonts}       
\usepackage{nicefrac}       
\usepackage{microtype}      
\usepackage{amssymb}
\usepackage{amsmath}
\usepackage{caption, subcaption}
\usepackage{multirow}
\usepackage{textcomp}
\usepackage{makecell}
\usepackage{colortbl}
\usepackage{pifont}
\usepackage{xspace}
\usepackage[ruled,noresetcount]{algorithm2e}
\usepackage[pagebackref=true,breaklinks=true,colorlinks,citecolor=citecolor,bookmarks=false]{hyperref}

\definecolor{citecolor}{RGB}{34,139,34}

\newcommand{\cmark}{\ding{51}\xspace}
\newcommand{\cmarkg}{\textcolor{lightgray}{\ding{51}}\xspace}
\newcommand{\xmark}{\ding{55}\xspace}

\makeatletter
\newcommand{\thickhline}{%
    \noalign {\ifnum 0=`}\fi \hrule height 0.5pt
    \futurelet \reserved@a \@xhline
}
\makeatother

\usepackage[capitalize]{cleveref}
\crefname{section}{Sec.}{Secs.}
\Crefname{section}{Section}{Sections}
\Crefname{table}{Table}{Tables}
\crefname{table}{Tab.}{Tabs.}

\definecolor{mypink}{RGB}{245,240,255}
\definecolor{mygray}{gray}{.9}
\definecolor{mycyan}{cmyk}{.1,0,0,0}
\def\sota{\textit{state-of-the-art} }

\usepackage{pifont}

\begin{document}
\title{Learning Self-Regularized Adversarial Views for Self-Supervised Vision Transformers}

\author{Tao~Tang$^\ast$,
        Changlin~Li$^\ast$,
        Guangrun~Wang,
        Kaicheng~Yu,
        Xiaojun~Chang,
        Xiaodan~Liang%
\IEEEcompsocitemizethanks{%
\IEEEcompsocthanksitem Tao. Tang is with Shenzhen Campus, Sun Yat-sen University, China. Email: trent.tangtao@gmail.com. Part of the work done when as an intern in DAMO Academy, Alibaba Group.
\IEEEcompsocthanksitem C. Li is with University of Technology Sydney, Australia. Email: changlinli.ai@gmail.com.
\IEEEcompsocthanksitem G. Wang is with Oxford University. Email: wanggrun@gmail.com.
\IEEEcompsocthanksitem K. Yu is with DAMO Academy, Alibaba Group. Email: kaicheng.yu.yt@gamil.com.
\IEEEcompsocthanksitem X. Chang is with University of Technology Sydney, Australia. Email: xiaojun.chang@uts.edu.au.
\IEEEcompsocthanksitem X. Liang is with Shenzhen Campus, Sun Yat-sen University, China. Email: xdliang328@gmail.com.
\IEEEcompsocthanksitem $^\ast$ Equal contribution.
}

\thanks{Corresponding author: Xiaodan Liang.}
}

\markboth{Work In Progress}%
{}

\IEEEtitleabstractindextext{%
\begin{abstract}
Automatic data augmentation (AutoAugment) strategies are indispensable in supervised data-efficient training protocols of vision transformers, and have led to state-of-the-art results in supervised learning.
Despite the success, its development and application on self-supervised vision transformers have been hindered by several barriers, including
the high search cost, the lack of supervision, and the unsuitable search space.
In this work, we propose \textbf{AutoView}, a self-regularized adversarial AutoAugment method, to learn views for self-supervised vision transformers, by addressing the above barriers.
First, we reduce the search cost of AutoView to nearly zero by learning views and network parameters simultaneously in a single forward-backward step, minimizing and maximizing the mutual information among different augmented views, respectively.
Then, to avoid information collapse caused by the lack of label supervision, we propose a self-regularized loss term to guarantee the information propagation.
Additionally, we present a curated augmentation policy search space for self-supervised learning, by modifying the generally used search space designed for supervised learning. 
On ImageNet, our AutoView achieves remarkable improvement over RandAug baseline (+10.2\% $k$-NN 
accuracy), and consistently outperforms \textit{sota} manually tuned view policy by a clear margin (up to +1.3\% $k$-NN accuracy). 
Extensive experiments show that AutoView pretraining also benefits downstream tasks  (+1.2\% mAcc on ADE20K Semantic Segmentation and +2.8\% mAP on revisited Oxford Image Retrieval benchmark)
and improves model robustness (+2.3\% Top-1 Acc on ImageNet-A and +1.0\% AUPR on ImageNet-O). Code and models will be available at \href{https://github.com/Trent-tangtao/AutoView}{https://github.com/Trent-tangtao/AutoView}.
\end{abstract}

}

\maketitle

\IEEEraisesectionheading{\section{Introduction}\label{sec:introduction}}
\IEEEPARstart{S}{elf-supervised} learning based on identifying augmented views of the data has made great progress in unsupervised visual representation learning \cite{bachman2019learning,chen2020simclr,he2020moco,grill2020byol,caron202dino}. These view-based self-supervised learning methods optimize the network parameters by contrastive learning \cite{chen2020simclr,he2020moco} or self-distillation \cite{grill2020byol,caron202dino} over different views of the same image. Recently, view-based self-supervised vision transformers (ViTs) have revealed emerging properties that have not been shown in either the supervised ViTs or previous unsupervised CNNs, attracting a lot of research interest in the community \cite{xie2021self,chen2021empirical,li2021esvit,zhou2021ibot,he2021mae}.

\begin{figure}[ht]
    \centering
\includegraphics[width=0.9\linewidth]{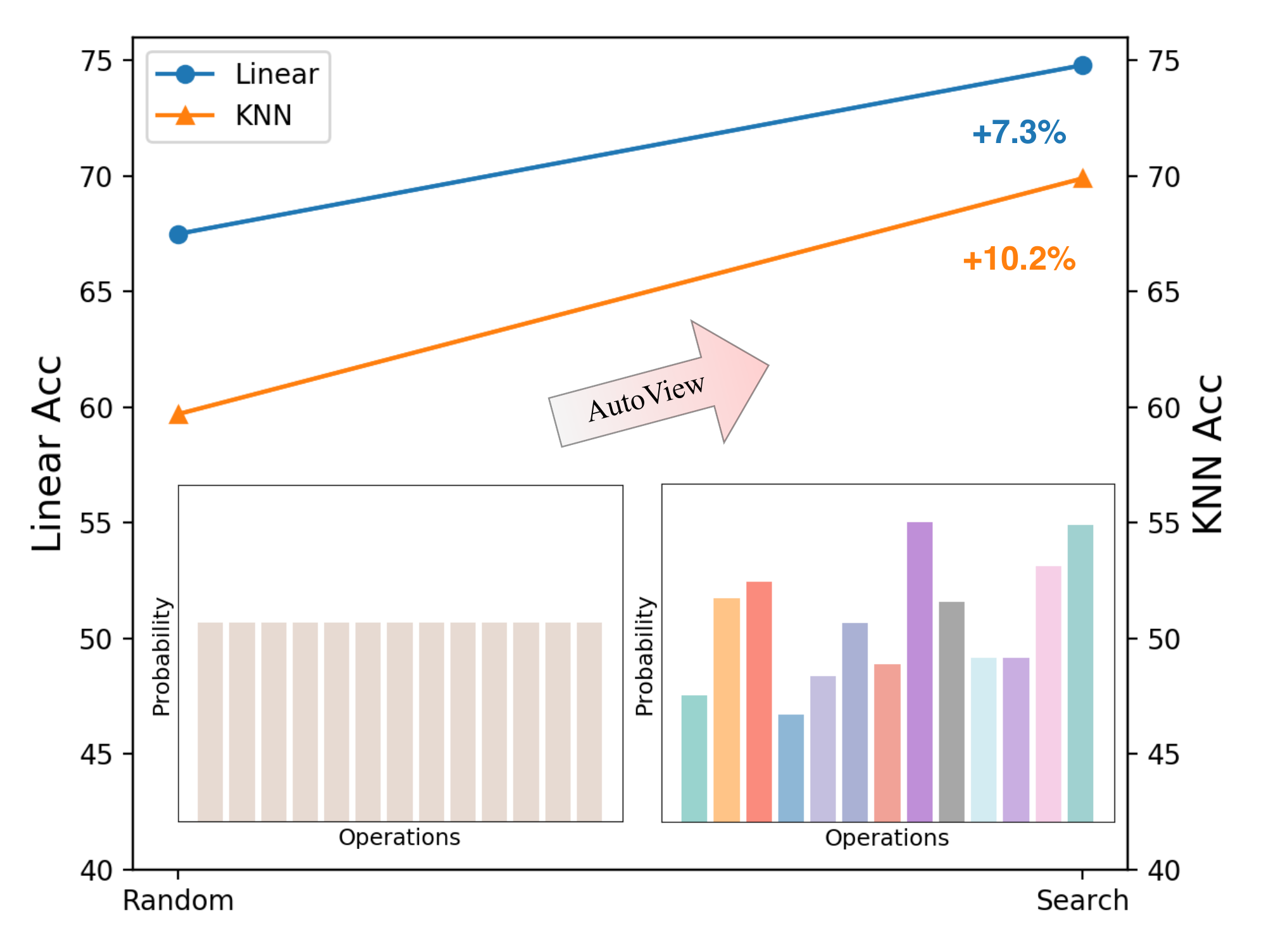}
    \caption{AutoView achieves remarkable improvement over RandAug baseline through searching operations' sampling probability.}
    \label{fig:intro}
\end{figure}

Augmentation policies are crucial for both supervised ViTs \cite{touvron2021deit} and view-based self-supervised learning \cite{chen2020simclr}. DeiT~\cite{touvron2021deit} improves the ImageNet accuracy of supervised ViT-L by 6.6\% by using better training recipes and better augmentation policies.
Meanwhile, \cite{chen2020simclr} show that data augmentation policies are crucial for learning good representations in view-based self-supervised learning.
However, designing augmentation policies of these views
requires considerable trial and error by human experts and could still be sub-optimal due to the limited design space. 

Recently, automatic data augmentation (AutoAugment) strategies have led to \sota results in supervised learning \cite{cubuk2019autoaugment,cubuk2020randaugment,li2020dada,liu2021ddas}.
These methods automatically search for improved data augmentation policies 
to achieve the highest validation accuracy on a target dataset. Among them, RandAug \cite{cubuk2020randaugment} has became an indispensable training recipe for supervised training of data-efficient ViTs.

Despite the success, there are still several barriers hindering its development and application in self-supervised learning.
%
First of all, AutoAugment methods are computationally expensive. As AutoAugment is a bi-level optimization problem, repeated training or iterative optimization of network parameters and augmentation policies are required, resulting in high search cost. Previous works on
efficient AutoAugment methods search augmentation policies on a subset of the dataset \cite{li2020dada,liu2021ddas,lim2019fastaug}. This proxy task is likely to be ineffective for augmentation search of self-supervised ViTs, as self-supervised ViTs rely on large datasets to learn properly.
Secondly, the lack of label supervision is another challenge for unsupervised AutoAugment methods. In supervised AutoAugment methods, augmentation policies are optimized by maximizing validation accuracy on a target dataset. Without label supervision, adversarial learning could be a useful way to learn augmentation policies \cite{zhang2019advaa,tian2020views,tamkin2020viewmaker,suresh2021adversarial}. However, without the constraint of label supervision as in semi-supervised adversarial view learning \cite{tian2020views}, these methods are prone to learning excessively strong augmentations, resulting in information collapse of the input.
Thirdly, the search space designed for supervised AutoAugment could be unsuitable for view-based
self-supervised learning. \cite{chen2020simclr} show that the sophisticated AutoAugment policies searched by supervised learning, do not work better than simple cropping + color distortion. Similarly, we empirically find that RandAug results in a remarkable performance drop in self-supervised ViTs.

In this work, we propose \textbf{AutoView}, a self-regularized adversarial AutoAugment method to learn views for self-supervised vision transformers, by addressing the aforementioned barriers. 
First, we propose to learn adversarial augmentation policies efficiently by learning views and network parameters simultaneously in a single forward-backward step. Specifically, augmentation policies are optimized by minimizing the mutual information among different augmented views, while network parameters are optimized by maximizing it. To avoid the information collapse or learning excessively strong augmentation policies due to the lack of label supervision, we propose a self-regularized loss term to guarantee that the useful information can be preserved by the augmentation. This loss term is designed to guarantee that the teacher models in view-based self-supervised learning methods can identify the augmented images as the original ones. 
Additionally, we present a curated augmentation policy search space for self-supervised learning, by removing harmful geometric transformations from the supervised AutoAugment search space and adding some augmentations that are generally used in manually designed view policies. The augmentation pipeline contains two sequentially applied randomly sampled augmentation operations with automatically optimized sampling weights, execution probability, and operation magnitude. Different from the previous search space, the execution probability of the first operation is set to 1 to eliminate the probability of not applying any augmentation. 

On ImageNet, our AutoView achieves remarkable improvement over RandAug baseline (+10.2\% $k$-NN and +7.3\% linear classification accuracy), as shown in \cref{fig:intro},  and outperforms \sota manually tuned view policy by a clear margin (up to +1.3\% $k$-NN accuracy). Extensive experiments show that AutoView pretraining also benefits downstream tasks (+2.8\% mAP on revisited Oxford Image Retrieval benchmark)
and improves model robustness (+2.3\% Top-1 Acc on ImageNet-A).
To validate the effectiveness of AutoView on new training schemes, we also conduct experiments on an efficient self-supervised learning scheme by progressively increasing the image size during training. 

Overall, our main contributions are three-fold:
\begin{itemize}
    \item We propose \textbf{AutoView}, a self-regularized adversarial AutoAugment method to learn views for self-supervised vision transformers, which addresses high learning cost and information collapse problem efficiently.
    \item We present a curated augmentation policy search space for self-supervised learning, in which random baselines are able to perform comparably to \sota manually-tuned policies.
    \item AutoView achieves remarkable improvements over RandAug baseline on $k$-NN and linear classification (\textbf{+10.2\%} and \textbf{+7.3\%}) and consistently outperforms hand-crafted policies on various~tasks.%
\end{itemize}

\section{Related Work}\label{sec:related_work}
\subsection{Self-supervised Visual Representation Learning}
Currently, a large body of works on self-supervised learning focus on discriminative approaches \cite{bachman2019learning,chen2020simclr,he2020moco,grill2020byol,caron202dino, arora2019theoretical, chen2021intriguing,hua2021feature,purushwalkam2020demystifying,wang2020understanding, xiao2020should,zbontar2021barlow,xie2021detco,xiong2020loco,lang2021contrastive}, which regard each image as a different class and trains the model by discriminating them up to data augmentations. From a mutual-information perspective \cite{wu2020mutual, tian2020views, bachman2019learning}, models are trained to maximize the mutual information between different augmented views of an image.
Plenty of positive and negative samples are needed to discover the similarity and dissimilarity. In practice, this requires large batches or memory banks. More recent works \cite{grill2020byol, chen2021simsiam, tian2021understanding} have shown that we can learn high-quality representations without negative samples. 

\textbf{Self-supervised Vision Transformers.}
MoCov3 \cite{chen2021empirical} first explored the iceberg of transformer-based self-supervised learning and presented a training recipe to let ViT perform reasonably well on ImageNet-1K linear evaluation.
MoBY \cite{xie2021self} adopts Swin Transformer \cite{liu2021swin} as the backbone to evaluate the learned representations on downstream tasks such as object detection and semantic segmentation.
DINO \cite{caron202dino} presents a new self-supervised learning method that shows good synergy with the transformer architecture and achieves the comparable performance of large self-supervised ConvNets using small/medium-size Transformers. Based on DINO, EsViT \cite{li2021esvit} further pursues efficient solutions to self-supervised vision transformers with two major insights: a multi-stage transformer architecture with sparse self-attentions, and a region-matching pre-training task. Concurrent BERT\cite{devlin2018bert} like approachs (BEiT\cite{beit}, MAE\cite{he2021mae}, CAE\cite{chen2022context}) propose to reconstruct raw pixels via mask image modeling.
And \cite{kong2022understanding} demonstrates that them also implicitly learns occlusion-invariant features, which are analogous to other discriminative methods, while the latter learns other invariance.
iBOT \cite{zhou2021ibot} performs masked image modeling via self-distillation as DINO with an online tokenizer.
Different from these works on objective, our work aims at developing an automatic view learning method rather than using hand-crafted view policies.


\textbf{Generation-based Pretext Tasks.}
At present, only a very small amount of work has explored automatic view generation for self-supervised learning on small-scale datasets preliminarily.
InfoMin \cite{tian2020views} leverages an adversarial semi-supervised view learning strategy. Except the flow-based view generation model which is adversarially trained to minimize mutual information, they need two classifiers on each of the learned views to perform classification using labels for the downstream task during the view learning process. InfoMin needs to first train the generative network with labels, then the frozen generator is used to generate views for self-supervised representation learning.
Viewmaker \cite{tamkin2020viewmaker} apply adversarial robustness literature to generate a variety of residual perturbations for a single input, with an iterative optimized image-to-image neural network as their Viewmaker network, requiring optimized in alternating steps with encoder. 
Their ineffectiveness prevents them from applying to self-supervised vision transformers that already require a large amount of computing resources.
In this work, we employ a constrained adversarial training method that 
adaptively reduces the mutual information between different views while preserving useful input features for the encoder to learn from just through one-step optimization.

\subsection{Automatic Augmentation}

Inspired by recent advancements in neural architecture search (NAS) \cite{zhang2020differentiable,liu2018darts}, researchers try to learn augmentation policies from data automatically \cite{cubuk2019autoaugment,zhang2019advaa,cubuk2020randaugment,li2020dada,liu2021ddas, suzuki2022teachaugment}.
AutoAugment \cite{cubuk2019autoaugment} first adopts reinforcement learning for auto-augmentation tasks but it requires searching for thousands of GPU-days. AdvAA \cite{zhang2019advaa} proposes online search manners by jointly optimizing augmentation policies and training the target networks. RandAug \cite{cubuk2020randaugment} directly uses naive grid search to find the best policy.
DDAS \cite{liu2021ddas} completes differentiable augmentations search through meta-learning with one-step gradient update. 
DADA \cite{li2020dada} introduces trainable policy parameters and adopts the Gumbel technique to update it with network weights by gradient descent algorithm alternately. 
These methods are based on supervised search space and need target validation sets, either have large search costs for multi-step optimization or multiple experiments, which are not suitable for self-supervised learning.
Thus this work proposes a self-regularized adversarial AutoView that can be optimized with self-supervised networks simultaneously in each single forward-backward step. 



\section{AutoView}\label{sec:methodology1}
\begin{figure*}[t]
    \centering
    \includegraphics[width=0.9\linewidth]{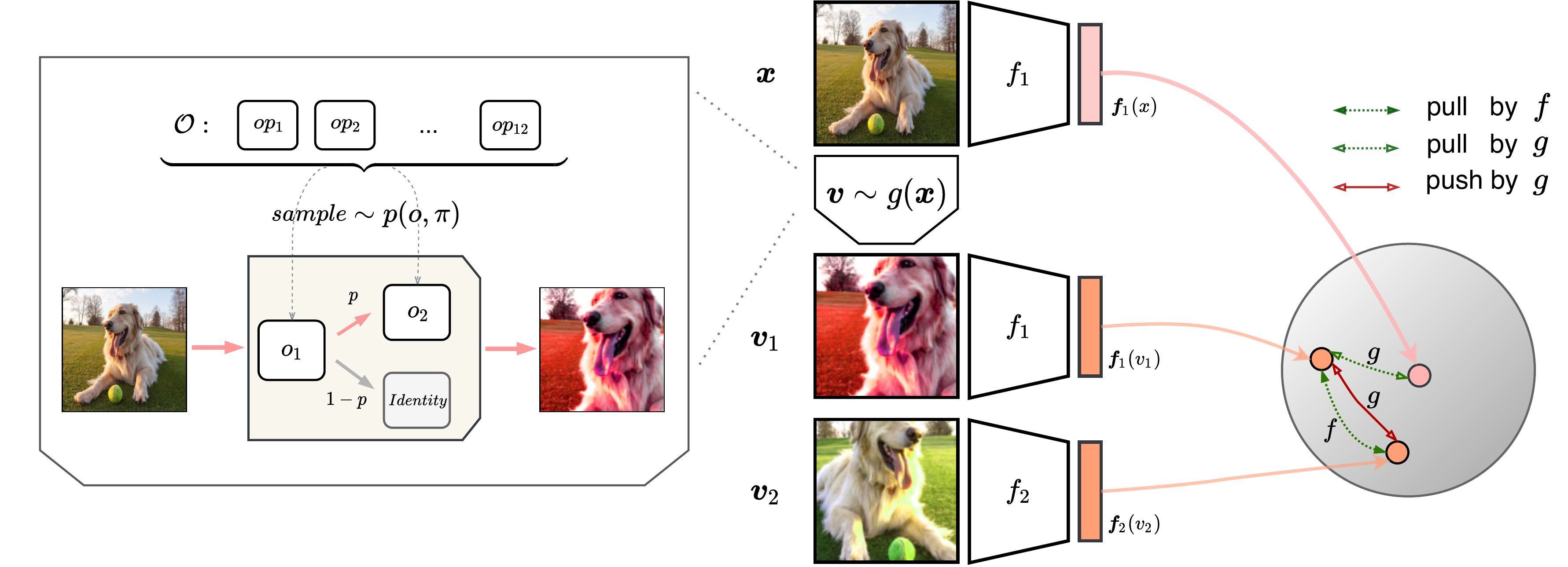}
    \caption{Adversarial AutoView Learning. Two encoders $f_1,f_2$ are trained to maximize $I(\bm{v_1}; \bm{v_2})$. The generator $g$ is adversarially trained to minimize $I (\bm{v_1}; \bm{v_2})$ while preserving useful information with $\bm x$. And the augmentation pipeline (left part) contains two operations sequentially sampled from $\mathcal{O}$. The execution probability of the first operation is set to 1 to avoid the situation without applying any augmentation.}
    \label{fig:autoview}
\end{figure*}

In this section, we first briefly formulate the standard view-based self-supervised vision transformers. And then we present our AutoView in detail, along with its three key elements: the one-step optimization (\cref{subsec:one-step}), the self-regularized information propagation (\cref{subsec:self-loss}), and the curated search space (\cref{subsec:ss}).
A visual illustration is in \cref{fig:autoview}.

\subsection{Learning Adversarial Views Through One-Step Optimization}
\label{subsec:one-step}


\textbf{View-based Self-supervised Vision Transformer.}
View-based self-supervised vision transformers have shown remarkable results in representation learning recently, whereby one image $\bm x$ is augmented using two separate data augmentations to get augmented views $\bm{v}_1, \bm{v}_2$, and then tranformers are trained to maximize the mutual information between different views i.e., $I(\bm v_1; \bm v_2)$, through maximize $\textit{sim}(f_1(\bm v_1), f_2(\bm v_2) )$, where $\textit{sim}(\cdot,\cdot)$ is a similarity function and $f_1, f_2$ are transformer encoder network. And $f_1$ parameters are updated with an exponential moving average (ema) of the $f_2$ parameters. But generating views through careful construction of data augmentation strategies is still a significant limitation, as several barriers are hindering automatic view learning development and application on self-supervised vision transformers.


\textbf{Adversarial View Learning Through One-Step Optimization.} 
Without label supervision, adversarial learning could
be a useful way to learn augmentation policies for self-supervised vision transformer.
For an input image $\bm{x}$, the view generator is defined as $g$
. Given $\bm{v_1},\bm{v_2} = g(\bm x)$, and two encoders $f_1,f_2$ are trained to maximize $I(\bm{v_1}; \bm{v_2})$.
Meanwhile, $g$ is adversarially trained to minimize $I (\bm{v_1}; \bm{v_2})$. Then the min-max objective is:\begin{equation}
    \min_{g}\max_{f_1,f_2} I \big(\bm{v_1}; \bm{v_2}\big).
    \label{eq:MI}
\end{equation} 
However, the ineffectiveness of multi-step optimization of the encoder and generator prevents it from applying to self-supervised
vision transformers that already require a large amount of
computing resources.

Hence, we employ a constrained adversarial training method that enables the model to adaptively reduce the mutual information between different views just through one-step optimization. In practice, we follow \cite{caron202dino} to learn to match the encoder networks' output probability distributions by minimizing the cross-entropy loss  w.r.t. the parameters of the $f_2$ encoder network with the $f_1$ network fixed. And our $g$ generates views by sampling according to probability, then we have:\begin{equation}
     \max_{g}\min_{f_2} \mathbb{E}_{\bm v_1, \bm v_2\sim g(\bm x)} H\big(f_1(\bm{v_1}), f_2(\bm{v_2}) \big),
  \label{eq:minmax}                                     
\end{equation} where $H(a, b) = - a \log b$.
We relax the policy selection to be differentiable through Gumbel-Softmax reparameterization \cite{jang2016gumbel} to achieve differentiable optimization of our image transforming $g$, details in \cref{subsec:ss}. It is worth noting that we realize the one-step optimization of the generator and encoder through Gradient reversal layer (GRL)\cite{GRL}. Remarkably, through that, the search cost of AutoView is nearly zero as views and network parameters are simultaneously optimized in each single forward-backward step.
Through one-step optimization, AutoView addresses the high learning cost and successfully performs AutoAugment for self-supervised learning in large-scale datasets for the first time.

\subsection{Avoid Collapse via Self-Regularized Information Propagation}
\label{subsec:self-loss}
With the loss function in Eqn.~\eqref{eq:minmax}, AutoView intends to increase the sampling probability of those transformations that can generate samples with high training loss. By sampling such transformations, AutoView can pay more attention to more aggressive augmentation strategies and increase model robustness against difficult samples. However, blindly minimizing the mutual information, i.e., increasing the difficulty of samples may cause the \textit{augment ambiguity phenomenon} \cite{wei2020circumventing}: augmented images may be far away from the majority of clean images, which could cause the under-fitting of models and deteriorate the learning process. InfoMin designed a semi-supervised method using a supervised constraint guiding the generator to alleviate this problem, which is not suitable for self-supervised learning without labels.

The Information Bottleneck (IB) theory \cite{tishby2000IB} uses the information theoretic objective to constrain the mutual information between the input and the representation, which can help us to preserve information. As previous work \cite{hu2019IB2} shows that the information constraining objective in supervised setting has the same form as that of self-supervised setting except the target outputs. Therefore, we unify these two objectives by using $\bm t$ as the output of the downstream tasks. In supervised setting, $\bm t$ represents the label of input $\bm x$. In self-supervised setting, $\bm t$ represents the input $\bm x$ itself. This leads to the unified objective function linking the representation $\bm z$ of input $\bm x$ and target $\bm t$ as:\begin{equation}
        \max I(\bm{z},\bm{t}).
  \label{eq:ib}                                         
\end{equation}This unified objective describes a constraint with the goal of maximizing the mutual information between the representation $\bm z$ and the target $\bm t$. Hence, we propose a self-regularized loss to avoid information collapse caused by the lack of label supervision:\begin{equation}
     \min_{g} \mathbb{E}_{\bm v\sim g(\bm x)}H \big(f(\bm v), f(\bm x)\big ),
  \label{eq:teacherloss}    
\end{equation}where $\bm x$ is the original image, $\bm v \sim g(\bm x)$ is the augmented image. So this self-regularized loss is to keep the information $I(\bm v,\bm x)$ to encourage distinguishable image features that share relevant semantic information with the original data.

In summary of Eqn.~\eqref{eq:minmax} and Eqn.~\eqref{eq:teacherloss}, the training process of our AutoView is formulated as follow:\begin{equation}\label{eq:total_loss}\small
    \max_{g}\min_{f_2} \mathbb{E}_{\bm v_1, \bm v_2} \Big\{H\big(f_1(\bm v_1), f_2(\bm v_2) \big) - H \big(f_1(\bm v_1), f_1(\bm x)\big)\Big\},
\end{equation}where the second term of the loss function is actually Eqn.\eqref{eq:teacherloss}. We use $f_1$ to achieve our information constraint on $g$, which will not affect the training of $f_2$, and then accomplish the conversion of maximization and minimization by the negative sign. So that we integrate the self-regularized loss into our adversarial view learning framework without undermining its advantage of one-step optimization.

\subsection {The Curated Augmentation Policy Search Space.}
\label{subsec:ss}
The current search space designed for supervised AutoAugment is unsuitable for view-based self-supervised learning as analyzed previously. Meanwhile, existing view learning methods, both the generative Viewmaker model and flow-based view generation model of InfoMin are pixel-wise generators, which are not suitable for view-based self-supervised learning of embedding space and can't be applied to large scale datasets due to the limitations of the pixel-wise image generator.
Thus we present a curated augmentation policy search space and design our online policy search framework based it.

In detail, we construct our search space of image pre-processing operations with 12 candidates. We denote the set of operations as $\mathcal{O}$: \{AutoContrast, Invert, Equalize, Solarize, Posterize, Contrast, Color, Brightness, Sharpness, Hue, Grayscale, Gaussianblur\}, removing geometric transformation(e.g., ShearY, TranslateX) \cite{hendrycks2019augmix} and adding augmentations commonly used in self-supervised methods (e.g., Gaussianblur, Grayscale) \cite{he2020moco}. Here we intuitively list the augmentation operations in our curated search space:
\begin{table}[h]
	\centering
 \resizebox{\linewidth}{!}{
	\begin{tabular}{llllllll}
		$\bullet\;$ \texttt{AutoContrast} &$\bullet\;$ \texttt{Invert} &$\bullet\;$ \texttt{Equalize} &$\bullet\;$ \texttt{Solarize} &\\
		
		$\bullet\;$ \texttt{Posterize} &$\bullet\;$ \texttt{Contrast} &$\bullet\;$ \texttt{Color} &$\bullet\;$ \texttt{Brightness} &\\
		
		$\bullet\;$ \texttt{Sharpness} &$\bullet\;$ \texttt{Hue} & $\bullet\;$ \texttt{Grayscale}&$\bullet\;$ \texttt{Gaussianblur}\\
	\end{tabular}
	}
\end{table}

Each augmentation policy  $\varphi$ contains $K$ image pre-processing operations: $ \varphi = {\left \{o_{k}  \right \} }^{K}_{k=1} $, where $ o_{k}\in\mathcal{O}$.
With $|\cdot|$ represents the number of elements in the set, we define sampling weights ${\bm \pi}\in\mathbb{R}^{K\times|\mathcal{O}|}$ to represent the sampling probability for each candidate operation, which is the parameter that the generator needs to learn. Therefore, the $k$-th image operation of policy can be sampled as:\begin{equation}\label{eq:softmax}
o_k \sim p(o;\bm {\pi}_k)=\mathrm{softmax}(\bm {\pi}_k)\in\mathbb{R}^{|\mathcal{O}|}.
\end{equation}For $i$-th policy, each image operation $o^i_k$ owns two parameters: the magnitude $m^i_k$ that determines the strength of the transformation and the probability of applying the transformation, $p^i_k$. So that the output of image $\bm x$ pre-processing $\tilde{o}^i_k$ can be computed as follows:\begin{align}\label{eq:op}
\begin{split}
    \tilde{o}^i_k(\bm x) = \left\{
        \begin{array}{ll}
        o^i_k(\bm x; m^i_{k}), &\text{with probability $ p^i_k$}  \\
        \bm x, &\text{with probability $1- p^i_k$}
        \end{array}.
    \right.
\end{split}
\end{align}


And it is worth noting that if the policy is not applied, that is when there is no augmentation, the self-supervised training will easily collapse. So instead of sampling and applying the whole policy by probability as in previous work \cite{li2020dada}, as shown in the left part of \cref{fig:autoview}, we sample the policy by layers and the execution probability of the first layer is fixed at 1. Finish formulating our search space and then we relax the policy selection to be differentiable through Gumbel-Softmax reparameterization \cite{jang2016gumbel}.

As to search adversarial policy online with encoder, we use the Gumbel-Softmax reparameterization trick \cite{jang2016gumbel} to achieve the differentiable relaxation. Thus, we regard Eqn.~\eqref{eq:op} as $ bO(\bm x; m)+(1-b)\bm x$, where $b \in\{0, 1\}$ is sampled from Bernoulli distribution. Since this distribution is non-differentiable, we instead use Relaxed Bernoulli distribution: \begin{equation}
    \mathrm{ReBern}(b; p, \lambda)=\sigma(\frac{1}{\lambda}\{\log\frac{p}{1-p}+\log\frac{u}{1-u}\}),
\end{equation}where $u$ is sampled from a uniform distribution on $[0, 1]$. 
Using this reparameterization, each operation can be differentiable w.r.t. its probability parameter $p$. For policy sample, we approximate their gradient in a similar manner to the straight-through estimator. For forward, we first sample operation by Eqn.~\eqref{eq:softmax}, e.g.$i$-th operation. Then:\begin{equation}
    \tilde{\bm c} = StopGrad \left(\bm h - \bm c \right) + \bm c,
\end{equation}where $\bm c = \mathrm{softmax}(\bm {\pi}_k)$ and $\bm h = one\_hot(\bm c,\bm i)$. So the backward pass uses differentiable variables $\bm c$ and the forward pass uses discrete variables $\bm h$.

Through the above relaxation, we can achieve differentiable optimization of image transforming $g$. So then the search cost of AutoView is nearly zero as we can learn views and network parameters simultaneously in Eqn.~\eqref{eq:total_loss}.



\section{Experiments}\label{sec:experiments}
This section presents our experimental setups and results. We first validate our AutoView with the standard self-supervised benchmark (\cref{subsec:classification_results}). And we also validate the effectiveness of AutoView on new progressive learning schemes (\cref{subsec:p_learn}). We then study the properties of the resulting features for segmentation, retrieval, transfer-learning, robustness and overfitting analysis (\cref{subsec:downstream}). Finally, we give the ablation study on the crucial components of AutoView (\cref{subsec:ablation}).

\subsection{Implementations Details}
\label{sec:implement}
\textbf{Architectures.}
We use the Vision Transformers with different amounts of parameters, ViT-S/16, ViT-B/16 as the encoder network $f$. 
For ViTs, /16 denotes the patch size being 16. We pretrain and finetune the transformers with 224-size images, so the total number of patch tokens is 196. The projection head $h$ is a 3-layer MLPs with $l_2$-normalized bottleneck following DINO\cite{caron202dino}.

\textbf{Pretraining Details.}
We pretrain the models on the ImageNet or OpenImages without labels. For self-supervised training, the hyperparameters follow closely to DINO. We use the adamw optimizer with a cosine decay learning rate schedule. The learning rate is linearly ramped up during the first 10 epochs to its base value (0.0005 for the batchsize of 256). The weight decay also follows a cosine schedule from 0.04 to 0.4. 

For AutoView policy training, we made a simplification on our preliminary search space, we discretized the continuous magnitude of augmentation and considered the same augmentation with different magnitudes as multiple operations. The operation number in an augmentation policy is set to $K=2$ and the two operations share sampling weight $\bm \pi$. Then we adopt Adam optimizer with $\beta = (0.5, 0.999)$ optimizing the policy weights $\pi$ with a step decay learning rate schedule, and the initial learning rate is 6e-5. The execution probability $p$ also optimized by Adam optimizer with $\beta = (0.5, 0.999)$ with learning rate 1e-5. 

For progressive learning, we divide the training process into four stages with about 25 epochs per stage: the early stage uses a small image size, while the later stages use larger image sizes with stronger regularization. And the minimum (for the first stage) and maximum (for the last stage) values of image size are 128, 224. The small view size of multi-crop training is from 55 to 96. Both of them change linearly according to the training stage.

For grid search of RandAug, RandAug makes the following simplifying assumptions: all operations share a single, discrete magnitude, $M \in \left [  0, 10\right ] $; all policies apply the same number of operations, $N$; all operations are applied with uniform probability. We select the best $k$-NN result from a grid search over $\left ( N,M   \right ) $.

\textbf{Linear and $k$-NN Evaluation Details.}
We follow the evaluation protocols in DINO. For linear evaluation, we sweep over different learning rates. For $k$-NN evaluation, we sweep over different numbers of nearest neighbors.

\textbf{End-to-end fine-tuning Details.}
We follow the protocol used in DeiT\cite{touvron2021deit} and finetune the features on downstream transfer tasks, e.g., Cifar10 and Flowers.


\textbf{Fine-Tuning Setting of Semantic Segmentation on ADE20K.} 
For semantic segmentation, we follow the configurations in iBOT \cite{zhou2021ibot}, fine-tuning $160$k iterations with $512\times512$ images and a layer decay rate of $0.65$ and set the learning rate on $8e^{-5}$.
We do not use multi-scale training and testing. 
To produce hierarchical feature maps, we use the features output from layer $4$, $6$, $8$, and $12$, with additional deconvolution layers appended after. 

\subsection{Main Classification Results}
\label{subsec:classification_results}
We evaluate our method on two large scale image datasets, ImageNet-1K \cite{deng2009imagenet} and OpenImages \cite{kuznetsova2018open}. ImageNet contains 1.2M train set images and 50K val set images in 1,000 classes. OpenImages consists of a total of 9.1 million images. We follow \cite{mishra2021object} to construct a subset of the OpenImages dataset that has 212K images present across 208 classes.

\begin{figure}[ht]
    \footnotesize
    \centering
    \includegraphics[width=0.9\linewidth]{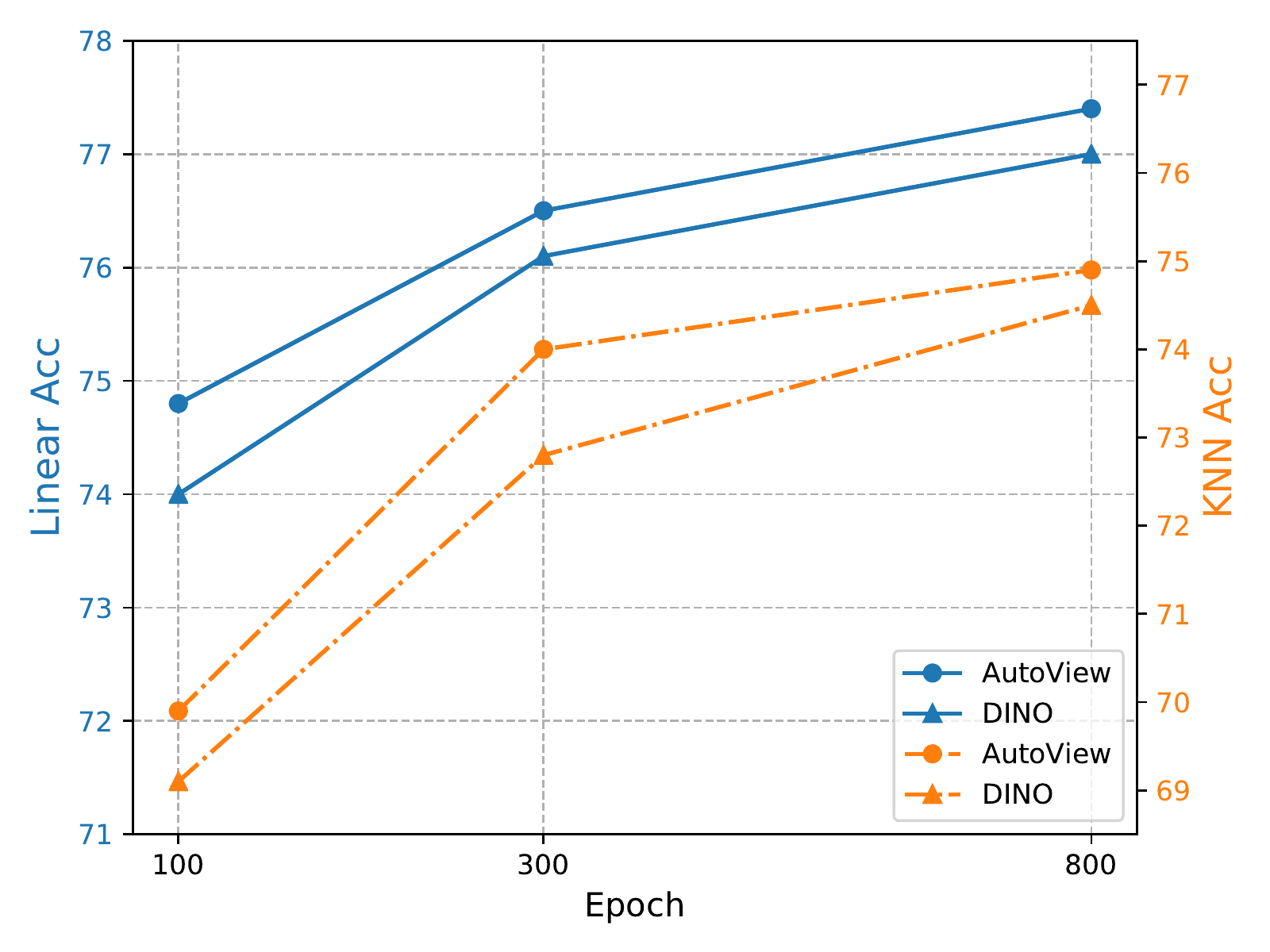}
    \caption{AutoView's improvements on different training lengths.}
    \label{fig:improvements}
\end{figure}

\begin{table}
    \footnotesize
    \centering
\resizebox{\linewidth}{!}{
    \begin{tabular}{l|cccc}
    \toprule
    Method & Augmentation & Arch  & Linear  & $k$-NN \\\midrule
      
   \multicolumn{5}{l}{\textbf{\textit{100 epochs}}} \\\midrule
    \multirow{4}{*}{MoCoV3\cite{chen2021empirical}} & Manual & ViT-S/16  & 67.0 & -  \\
  &\cellcolor{mypink}  AutoView & \cellcolor{mypink} ViT-S/16 &\cellcolor{mypink}  \textbf{67.5} & \cellcolor{mypink}  - \\
  
 &Manual& ResNet-50 & 68.9 & -  \\
 &\cellcolor{mypink}  AutoView & \cellcolor{mypink} ResNet-50 & \cellcolor{mypink}  \textbf{69.3} & \cellcolor{mypink} -  \\
 \arrayrulecolor{lightgray}\hline \arrayrulecolor{black}

 \multirow{2}{*}{iBOT\cite{zhou2021ibot}} & Manual & ViT-S/16  & 74.4 & 70.7 \\
  & \cellcolor{mypink}  AutoView & \cellcolor{mypink} ViT-S/16 & \cellcolor{mypink} \textbf{74.8} & \cellcolor{mypink} \textbf{71.2} \\
   \arrayrulecolor{lightgray}\hline\arrayrulecolor{black}
   
      \multirow{3}{*}{DINO\cite{caron202dino}} & Manual& ViT-S/16  &74.0& 69.1\\
    & RandAug & ViT-S/16 & 67.5&59.7\\
&\cellcolor{mypink}   AutoView & \cellcolor{mypink} ViT-S/16  &\cellcolor{mypink} \textbf{74.8} &\cellcolor{mypink} \textbf{69.9} \\
\arrayrulecolor{lightgray}\hline\arrayrulecolor{black}
  \multirow{2}{*}{DINO*\cite{caron202dino}} & Manual &  ViT-S/16 &73.7& 69.2\\
 & \cellcolor{mypink}  AutoView &\cellcolor{mypink} ViT-S/16 &\cellcolor{mypink}  \textbf{74.5} & \cellcolor{mypink} \textbf{70.5} \\

\midrule

\multicolumn{5}{l}{\textbf{\textit{300 epochs}}} \\\midrule
   \multirow{2}{*} {DINO\cite{caron202dino}} & Manual & ViT-S/16& 76.1& 72.8 \\
 &  \cellcolor{mypink}  AutoView & \cellcolor{mypink} ViT-S/16& \cellcolor{mypink} \textbf{76.5} & \cellcolor{mypink} \textbf{74.0} \\
\midrule
\multicolumn{5}{l}{\textbf{\textit{400 epochs}}} \\
\midrule
\multirow{2}{*}{DINO\cite{caron202dino}}&Manual& ViT-B/16&78.2 & 76.1 \\
& \cellcolor{mypink}  AutoView & \cellcolor{mypink} ViT-B/16&\cellcolor{mypink}  \textbf{78.6} & \cellcolor{mypink} \textbf{76.5} \\
\midrule
\multicolumn{5}{l}{\textbf{\textit{800 epochs}}} \\
\midrule
\multirow{2}{*}{DINO\cite{caron202dino}}&Manual& ViT-S/16& 77.0& 74.5 \\
& \cellcolor{mypink} AutoView &\cellcolor{mypink}  ViT-S/16& \cellcolor{mypink} \textbf{77.4} & \cellcolor{mypink} \textbf{74.9} \\
 \bottomrule
\end{tabular}
}
\caption{Linear and $k$-NN classification on ImageNet. * denotes progressive training. Our method is highlighted in \colorbox{mypink}{purple}.}
    \label{tab:Imagenet}
\end{table}

\textbf{Linear and $k$-NN results on ImageNet.} 
To evaluate the quality of pre-trained features, we either use a linear classifier on the frozen representation or a $k$-nearest neighbor classifier ($k$-NN) . 
In \cref{tab:Imagenet}, AutoView achieves remarkable improvements over RandAug baseline on $k$-NN and linear classification (+10.2\% and +7.3\%) and steadily boosts both the top-1 and $k$-NN classification accuracy on ImageNet for different frameworks (e.g. DINO, iBOT, MoCoV3)
, different architectures (e.g.ViT-S,ViT-B) and different training lengths (e.g.100$\sim$800 epochs). 
To be specific, AutoView can promote the $k$-NN accuracy of DINO with ViT-S of various train periods by 0.8\%, 1.2\%, 0.4\% respectively, and the results are shown in \cref{fig:improvements} intuitively show our continuous boost on different training lengths(e.g. 100$\sim$800 epochs). Particularly, AutoView consistently boosts the DINO on the progressive training setting with the gain of 1.3\% $k$-NN accuracy.

\textbf{Why we focusing on ViTs?}
First, data augmentation is more critical in ViTs than CNNs. The vanilla ViT needs to be trained with a large dataset with full supervision (e.g., ImageNet-21k). DeiT improves the vanilla ViT-L by up to 2.2\% using better augmentation policies proposed by RangAugment. Second, more and more current sota self-supervised methods use ViTs as their backbone, and our method is in line with them.
Moreover, our AutoView is a general AutoAugment method for self-supervised learning. As shown in \cref{tab:Imagenet}, our method consistently improves the performance when applied to CNN-based self-supervised learning approach.

\begin{table}[ht]
    \footnotesize
    \centering
 \resizebox{0.96\linewidth}{!}{
    \begin{tabular}{lcccc}
    \toprule
     Method & Augmentation& Arch  & Epoch  & mAP \\\midrule
     MoCov2$\dagger$ & - & ResNet-50 & 200 & 58.6 \\
     DINO & Manual& ViT-S/16 & 100 & 59.6\\
\rowcolor{mypink}DINO &  AutoView & ViT-S/16 &100 & \textbf{60.3}\\
 \bottomrule
    \end{tabular}
    }
    \caption{Classification on OpenImage. $\dagger$ denotes using object-aware crops.}
    \label{tab:openimages}
\end{table}

\textbf{Classification results on OpenImages.} 
We firstly pretrain models on OpenImages dataset. After pre-training, we freeze the backbone weights and train a linear classifier with a multi-class logistic regression loss. We follow the mAP metric as described in \cite{mishra2021object}. The last row in \cref{tab:openimages} shows the continuous boost obtained by using our AutoView on OpenImages.

\subsection{Progressive Learning for Self-supervised training}
\label{subsec:p_learn}
As self-supervised training is time-consuming and image size plays an important role in training efficiency. 
\cite{yu2019pda,press2020shortformer,karras2017progressivegan,tan2021efficientnetv2} have proposed different kinds of progressive training, which dynamically change the training settings or networks.
We take the insight of EfficientNetV2\cite{tan2021efficientnetv2} and present an efficient self-supervised learning scheme by progressively increasing the image size during training. We divide the training process into four stages. In the early training epochs, we trained the network with smaller images such that the network can learn simple representations easily and fast and then gradually increase image size. As the result shown in \cref{tab:prog}, we indeed improve the training speed but it also comes with a drop in accuracy. In this section, all models are trained for 100 epochs with ViT-S/16 and train time is counted on the same machine using 8 RTX 2080Ti GPUs with the total batch size 256.

\begin{table}[ht]
    \footnotesize
    \centering
 \resizebox{\linewidth}{!}{
    \begin{tabular}{lccc|c}
        \toprule
    Method & Train Scheme &Linear  & $k$-NN & Time \\\midrule
    DINO & Manual& 74.0 & 69.1 & 61h\\
    \midrule
    DINO &Progressive &  73.7(-0.4)& 69.2(+0.1) &  \\
    DINO & Progressive RandAug &  74.2(+0.1)& 70.1(+1.0)& 46h \\
\rowcolor{mypink} DINO & Progressive AutoView & 74.5(+0.5) & 70.5(+1.5) &\cellcolor{white} \\
        \bottomrule
    \end{tabular}
}
\caption{Training time and results on ImageNet of different train schemes.}
    \label{tab:prog}
\end{table}

\textbf{Progressive Learning with AutoView.} As EfficientNetV2 points out using the same regularization for all image sizes causes the drop in accuracy. To improve both training speed and accuracy, they also adaptively adjust regularization linearly according to image size. To continue to explore this property, we adapt grid search over $(N,M)$ as RandAug for DINO progressive training in our search space to select the best result of linear classification on imagenet.

With the search result $\{(2,3),(2,6),(2,6),(2,8)\}$ of four train stage, we improve both the accuracy and training time. 
The above results prove that adaptive regularization helps progressive learning gain increase in accuracy. But the RandAug search results also show that simple linear schedule regularization just according to image size is not always the most suitable. 
Then thanks to AutoView, we use it to dynamically adjust augmentations during model training and get better results than costly and tedious manual search, which clearly demonstrates the effectiveness and generalization of AutoView.

\subsection{Downstream Tasks}
\label{subsec:downstream}
We first evaluate our method in the dense downstream tasks, semantic segmentation, on which we observe improvements over the vanilla pre-trained baselines.
We then evaluate properties of the learned features of models pretrained for 300 epochs with ViT-S/16 in terms of nearest neighbor search and transferability to different datasets.  


\begin{table}[ht]
\footnotesize
\centering
 \resizebox{\linewidth}{!}{
\begin{tabular}{lccccc}
\toprule 
Method &Augmentation & Arch & mIoU  & aAcc & mAcc\\
\toprule

\textcolor{gray!80}{Sup.} &- & ViTB/16 & \textcolor{gray!80}{46.6} &-&-\\
DINO & Manual& ViTB/16 & 46.8  & 82.4 & 56.6\\
\rowcolor{mypink} DINO & AutoView & ViTB/16 & \textbf{47.1} & \textbf{82.8} & \textbf{57.8}\\
\bottomrule
\end{tabular}}
\caption{Semantic segmentation on ADE20K.}
\label{tab:ade}
\end{table}

\textbf{Semantic Segmentation.} 
We evaluate our method in the dense downstream task, semantic segmentation, on which we observe improvements over the vanilla pretrained baseline. And in this experiment models are pretrained for 400 epochs with ViT-B/16. Our semantic segmentation experiments on ADE20K\cite{zhou2017ADE} use UperNet \cite{xiao2018up} following the code in iBOT. We use the task layer in UperNet and fine-tune the entire network. From \cref{tab:ade}, we can see that AutoView advances DINO with ViT-B/16 with a clear margin of 1.2\% on mAcc.

\begin{table}[ht]
    \footnotesize
    \centering
\resizebox{\linewidth}{!}{
    \begin{tabular}{lccccc}
        \toprule
      \multirow{2}{*}{Method} & \multicolumn{3}{c}{Video object segmentation}&  \multicolumn{2}{c}{$\mathcal{R}$Oxford} \\
      \cmidrule(lr){2-4} \cmidrule(lr){5-6} & $(\mathcal{J}$\&$\mathcal{J})_m$ & $\mathcal{F}_m$ & $\mathcal{F}_m$ & Medium & Hard \\\midrule
        DINO Manual&  61.3    & 62.9   &  59.7 &  33.7& 11.9  \\
\rowcolor{mypink}DINO  AutoView & \textbf{61.7}    & \textbf{63.1} &  \textbf{60.3} & \textbf{36.5}  & \textbf{13.6} \\
        \bottomrule
    \end{tabular}
}
\caption{Video object segmentation and Image retrieval.}
    \label{tab:DAVIS and retrieval}
\end{table}

\textbf{Video Object Segmentation.} We evaluate the output patch tokens on the DAVIS-2017 video instance segmentation benchmark\cite{pont2017davis}. We segment scenes with the nearest neighbor between consecutive frames, while do not train any model on top of the features, nor fine-tune any weights for the task, and follow the experimental protocol as DINO. 
According to \cref{tab:DAVIS and retrieval}, AutoView is superior to vanilla DINO on all metrics.

\textbf{Image Retrieval.} 
We consider the revisited Oxford \cite{radenovic2018revisiting}, which contains 3 different splits of gradual difficulty with query/database pairs. We freeze the features and directly apply $k$-NN for retrieval as DINO, reporting the Mean Average Precision (mAP) for the Medium and Hard splits. Since AutoView has higher $k$-NN results on Imagenet-1K, the performance is also better for AutoView in the image retrieval task (up to 2.8\%) as in \cref{tab:DAVIS and retrieval}.

\begin{table}[ht]
\footnotesize
    \centering
\resizebox{1\linewidth}{!}{
    \begin{tabular}{lcccc}
        \toprule
      Method   & Cifar10 & Cifar100  &Flwrs& Cars  \\\midrule
\textcolor{gray!80}{Rand.} & \textcolor{gray!80}{99.0} & \textcolor{gray!80}{89.5}& \textcolor{gray!80}{98.2} & \textcolor{gray!80}{92.1}\\
 DINO Manual& 99.0 & 90.5 & 98.3 &92.6 \\
\rowcolor{mypink}DINO  AutoView &\textbf{99.0}  & \textbf{90.5} & \textbf{98.7} & \textbf{93.3}    \\
        \bottomrule
    \end{tabular}
}
\caption{Transfer learning on different datasets.}
\label{tab:transfer}
\end{table}

\textbf{Transfer learning.} We evaluate the quality of the features pretrained on ImageNet-1K and fine-tune on several smaller datasets. The results are demonstrated in \cref{tab:transfer}. While the results on CIFAR10 and CIFAR100 have almost plateaued, AutoView consistently boosts the baseline.

\subsection{Robustness and Overfitting Analysis}
AutoView enables models to cover more augmentation policies. This property could improve model robustness to novel examples. To verify if AutoView can improve ViT-based models’ robustness and out-of-distribution performance, we evaluated our pretrained models on two robustness scenarios including natural adversarial example and out-of-distribution detection. And we also verify that our method has no overfitting problem on ImageNetV2 \cite{recht2019imagenetv2}.
In this section, we study the robustness and overfitting with ViT-S/16 pretrained for 300 epochs and then linearly evaluated for 100 epochs.

\begin{table}[ht]
   \footnotesize
    \centering
 \resizebox{\linewidth}{!}{
    \begin{tabular}{lcccc}
    \toprule
\multirow{2}{*}{Method}  & \multicolumn{2}{c}{Nat. Adversarial Example} & \multicolumn{1}{c}{Out-of-Dist}& \multicolumn{1}{c}{IN-V2} \\ 
  \cmidrule(lr){2-3} \cmidrule(lr){4-4}  \cmidrule(lr){5-5}     & \multicolumn{1}{c}{Top1-Acc}  & \multicolumn{1}{c}{AURRA} & \multicolumn{1}{c}{AUPR} & \multicolumn{1}{c}{$k$-NN} \\ \midrule
  DINO Manual&  11.9   & 16.1& 21.6 & 61.4\\
\rowcolor{mypink}DINO AutoView &\textbf{13.2}   & \textbf{18.2} & \textbf{22.6} &  \textbf{62.3}\\

 \bottomrule
    \end{tabular}
}
 \caption{Model's robustness against natural adversarial and out-of-distribution examples and performance on ImageNetV2 (IN-V2).}
    \label{tab:robust}
\end{table}

\textbf{Natural Adversarial Example.} The dataset ImageNet-A \cite{hendrycks2021natural} adversarially collects 7500 unmodified, natural but “hard” real-world images which are drawn from some challenging scenarios (e.g., fog scene and occlusion). The metric for assessing classifiers’ robustness to adversarially filtered examples includes the top-1 accuracy
and Area Under the Response Rate Accuracy Curve (AURRA). 
AURRA is an uncertainty estimation metric introduced in \cite{hendrycks2021natural}. 
As shown in \cref{tab:robust}, the ViT trained by AutoView improves over the vanilla pretrained DINO by 2.1\% AURRA.

\textbf{Out-of-distribution Detection.} The dataset ImageNet-O\cite{hendrycks2021natural} is an adversarial out-of-distribution detection dataset which adversarially collects 2000 images from outside ImageNet-1K. The anomalies of unforeseen classes should result in low-confidence predictions. The metric is the area under the precision-recall curve (AUPR) . As \cref{tab:robust} indicates, AutoView outperforms 
baseline by 1.0\% AUPR.

\textbf{Performance on ImageNetV2.}
\cref{tab:robust} reports the additional evaluation on ImageNet V2\cite{recht2019imagenetv2}, that has a test set distinct from the ImageNet validation, which reduces overfitting on the validation set. The results (+0.9\% $k$-NN) verify the generalization and effectiveness of our method. Note that our method does not use the validation set for augmentation searching, so there should be no overfitting problem.

\subsection{Ablation study}
\label{subsec:ablation}
In this section, we perform extensive ablation studies to analyze each components of our proposed AutoView. All the results in this section are obtained by training ViT-S/16 with different schemes for 100 epochs.

\begin{figure}[ht]
    \centering
    \includegraphics[width=1\linewidth]{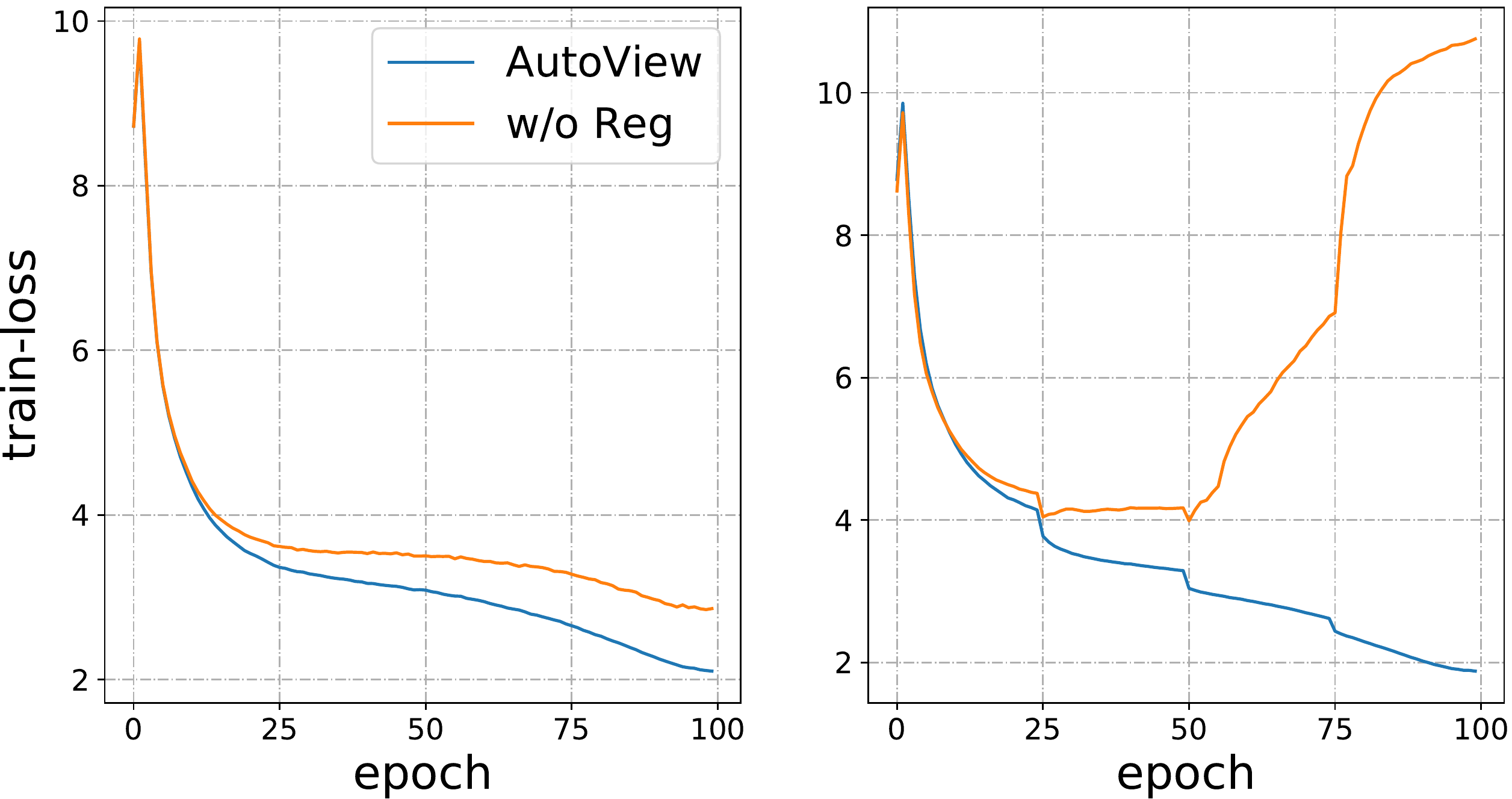}
    \caption{Learning curves of different learning schemes. The right one is on Progressive Learning.}
    \label{fig:reg_loss}
\end{figure}

\textbf{Effectiveness of the Self-Regularized Loss.}
We compare the pretraining objective with and without self-regularized loss. As shown in \cref{fig:reg_loss}, 
the training curve with our self-regularized loss is much stabler, and converges to a lower loss value, which demonstrate that the self-regularized loss can effectively avoid information clapse. As shown in the fourth column of \cref{tab:ablation}, without self-regularized loss (Reg. Loss), the performance dropped by 0.4\% in regular AutoView training and dropped considerably by 3.7\% in progressive learning, proving its importance in AutoView.

\begin{table}[ht]
    \footnotesize
    \centering
    \resizebox{\linewidth}{!}{
    \begin{tabular}{lccc}
    \toprule
       Method & SS &  Linear & $k$-NN \\\midrule
     DINO RandAug & RandAug & 67.5 & 59.7\\
 \rowcolor{mypink} \cellcolor{white}DINO RandAug & AutoView &  74.0(+6.5) & 69.3(+9.6) \\
     \midrule
     DINO AutoView & RandAug & 69.9  &63.5  \\
  \rowcolor{mypink} \cellcolor{white}DINO AutoView & AutoView &  74.8(+4.9) & 69.9(+6.4) \\
 \bottomrule
    \end{tabular}}
    \caption{Effectiveness of the curated augmentation policy Search Space (SS).}
\label{tab:ss}
\end{table}

\textbf{Effectiveness of the Search Space.}
For our search space, we compared our present curated augmentation policy search space with the one designed for supervised learning that is generally used by popular AutoAugment methods. 
As shown in \cref{tab:ss}, AutoView and RangAug both gain a large margin of linear and $k$-NN accuracy with our search space. For the first time, the AutoAugment method for self-supervised learning is proved to be able to perform in a policy search space.
For AutoView, we can find it significantly outperforms RandAug on same search space (up to +3.8\% $k$-NN accuracy). And it is worth mentioning that the results of RandAug are obtained through tedious and time-consuming grid search.

\begin{table}[ht]
    \centering
\resizebox{\linewidth}{!}{
    \begin{tabular}{c|c|c|c|c}
    \toprule
H. Policy &  Exc. Prob. &  Weight & Reg. Loss & $k$-NN \\\midrule
 \cmark & \cmark & \cmark & \cmark &\cellcolor{mypink} 69.9  \\
\colorbox{mycyan}{\xmark} & \cmarkg & \cmarkg & \cmarkg& 69.1(-0.8)\\  
\cmarkg & \colorbox{mycyan}{\xmark} & \cmarkg & \cmarkg & 68.3(-1.6)\\
 \cmarkg & \cmarkg & \colorbox{mycyan}{\xmark} &\cmarkg & 69.0(-0.9)\\ 
 \cmarkg & \cmarkg & \cmarkg & \colorbox{mycyan}{\xmark}& 69.5(-0.4) \\\midrule
  \cmark & \cmark & \cmark & \cmark & \cellcolor{mypink} 70.5  \\
\cmarkg & \cmarkg & \cmarkg &\colorbox{mycyan}{\xmark} & 66.8(-3.7)\\
\bottomrule
\end{tabular}
}
 \caption{Results on ImageNet of different train settings. The results of the last two lines are obtained in progressive learning.}
\label{tab:ablation}
\end{table}

\textbf{Importance of Searching.}
As shown in \cref{tab:ablation}, we further investigate the effect of searching for sampling weights and execution probability of AutoView. In the first column, we search policy hierarchically (H. Policy) instead of sampling and applying the whole policy by probability. Then we fix the execution probability at 0.5 without searching (Exc. Prob.). Next, we randomly select views in our policy search framework rather than sampling by the weights searched (Weight). The clear accuracy drop reflects the searching of hierarchical policy, execution probability and views are all vital components of AutoView.

\begin{table}[ht]
    \centering
\resizebox{\linewidth}{!}{
    \begin{tabular}{c|cccccc}
        \toprule
  $\alpha$ & 0 & 0.3&0.5&0.8&\cellcolor{mypink}1.0 &3.0  \\\midrule
  $k$NN-20ep  &
  47.41 & 47.53& 47.57& 47.64& \cellcolor{mypink}\textbf{47.72}& 47.52
  \\
        \bottomrule
    \end{tabular}
}
\caption{The trade-off of different weight settings. The evaluations are on progressive learning. }
\label{tab:ablation3}
\end{table}

\begin{figure}[ht]
    \centering
    \includegraphics[width=1\linewidth]{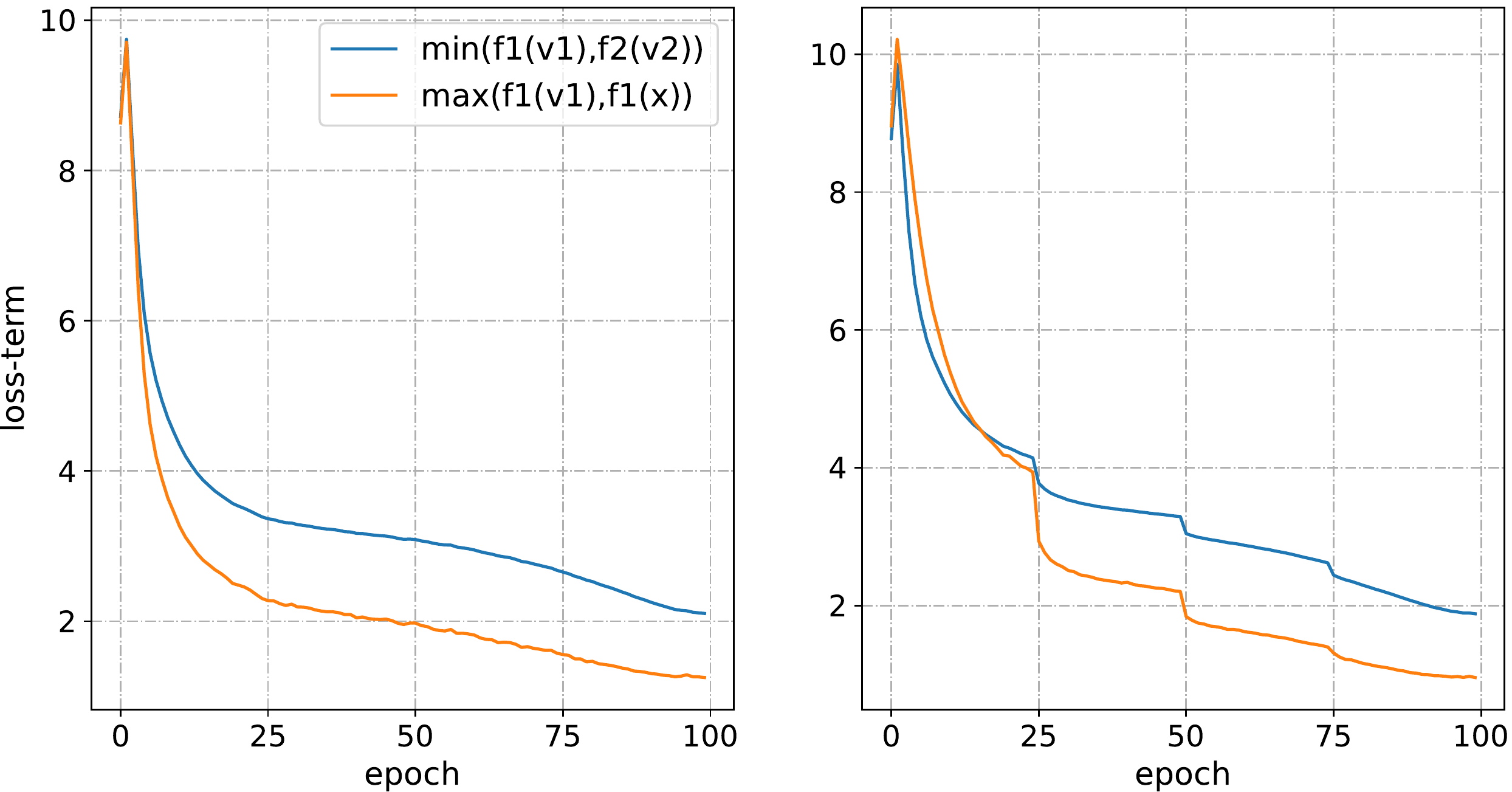}
    \caption{Learning curves of two objectives during training.}
    \label{fig:trade-off}
\end{figure}

\textbf{The trade-off of two objectives in AutoView.} There exists a trade-off between $min(f_1(v_1),f_2(v_2))$ and $max(f_1(v_1), f_1(x))$ in our AutoView. And we have considered and verified the trade-off between these two objectives through a loss weight $\alpha$, which multiplies on the self-regularized item. The results are shown in \cref{tab:ablation3}, where we can see that $\alpha$ is 1.0 is the optimal setting we adopt in this paper. 
Moreover, our self-regularized loss is the constraint of the augmented view and the original view on the single network $f_1$, which is much weaker than the loss between the two siamese networks and just can be seen as one regularization term, and the regularization term is adaptively adjusted during training as the \cref{fig:trade-off} shows. 

\begin{table}[ht]
    \centering
\resizebox{\linewidth}{!}{
    \begin{tabular}{c|ccccc}
        \toprule
  Setting  & \cellcolor{mypink}AutoView & Policy-retrain & SS-allop & $K=3$ &   Not-share $g$   \\\midrule
  $k$NN-10ep  & \cellcolor{mypink}\textbf{42.7} & \textbf{42.7}  & 41.1  & 40.6       & 41.1 \\
        \bottomrule
    \end{tabular}
}
\caption{Different settings of AutoView.}
\label{tab:ablation2}
\end{table}

\begin{figure*}[ht]
\centering
\includegraphics[width=1.0\linewidth]{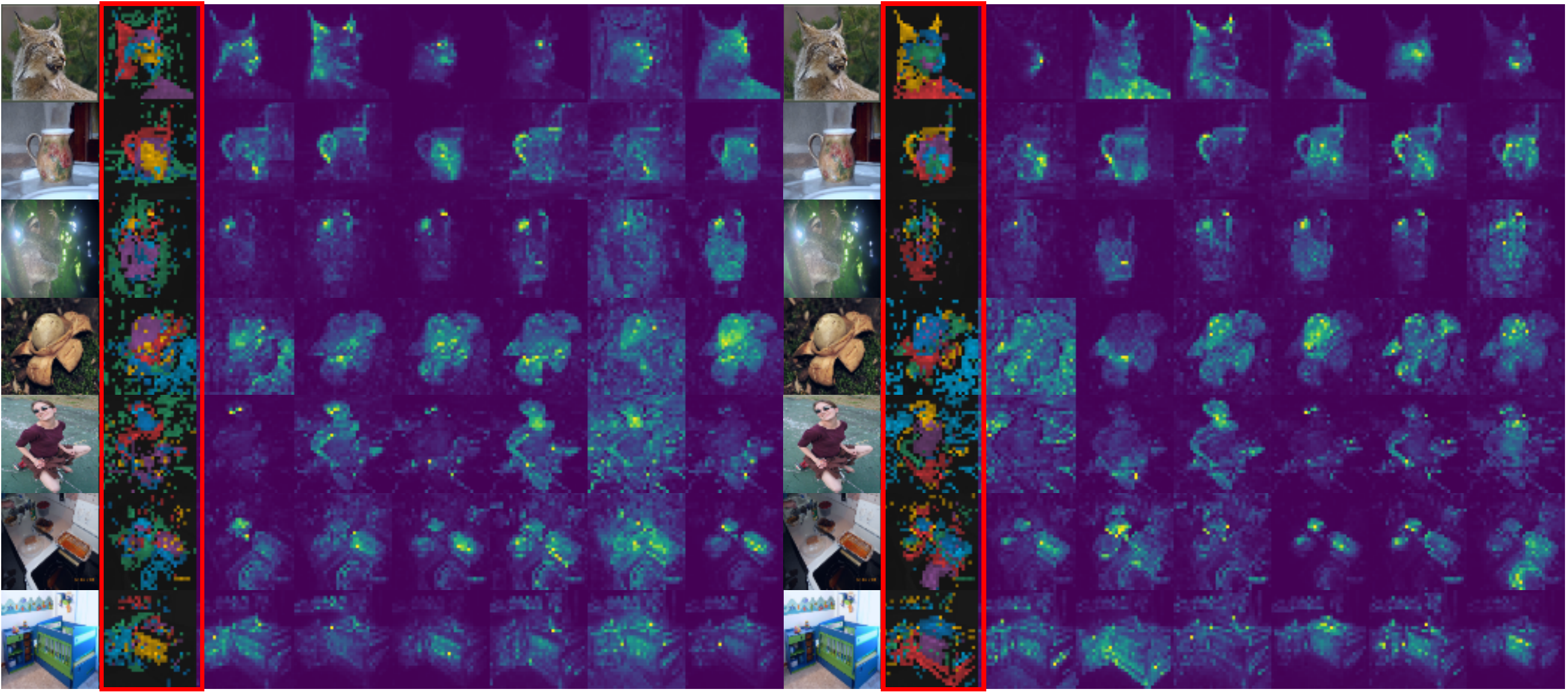}
\caption{Visualization for self-attention map from multiple heads of a ViT-S/16 trained for 300 epochs with DINO Manual (left) and our AutoView (right).
}
\label{fig:attnmap1}
\end{figure*}

\textbf{Method Complexity.} Our searched augmentation policy is general and universal. One can directly use our searched policy to get the same performance results without searching.
As shown in \cref{tab:ablation2}, the model retraining with searched policy (Policy-retrain in \cref{tab:ablation2}) performs comparable with the the model that use AutoView in an online way (AutoView in \cref{tab:ablation2}).

\textbf{Search Space Design.}
We adopt the supervised RA search space and make non-trivial modifications. The vanilla RA search space did not contain the transformations commonly used in SSL. We first added them, but then we found that the effect was unsatisfactory (see \cref{tab:ablation2}). With many geometric transformations combined, the object often shifts from the center of the image \cite{hendrycks2019augmix}, which hurts SSL model learning, so we removed these transformations.

\textbf{Choice of $K$.} First, the $K$ of \textit{sota} methods (i.e., AA, RA, DADA, etc.) is set to 2. Second, our method is based on RA. During RA searching (see \cref{subsec:p_learn}), we found that $K$=2 achieved the best results. We also explore different $K$s in \cref{tab:ablation2}. 

\textbf{Result when $g$ is not shared for $v_1$ and $v_2$.} We have considered and verified various cases of network structure design. When $g$ is not shared, the result is sub-optimal (see \cref{tab:ablation2}). 

\begin{table}[ht]
    \centering
\resizebox{\linewidth}{!}{
    \begin{tabular}{l|ccccc}
        \toprule
  Method      & \cellcolor{mypink}AutoView  & InfoMin & Viewmaker & DADA &   RandAugment\\\midrule
 Overhead  & \cellcolor{mypink}\textbf{0x}  & 1x & 1x & 1x & 4-80x \\
        \bottomrule
    \end{tabular}}
\caption{Search overhead of different methods.}
\label{tab:overhead}
\end{table}
\textbf{Search Overhead.} The search overhead of different methods in \cref{tab:overhead} shows the advantage of our one-step optimization. AutoView learns adversarial augmentation policies efficiently by learning views and network parameters simultaneously in a single forward-backward step without additional search costs. Noting that, the search overhead there is just a rough estimate, as for InforMin and Viewmaker, we did not consider the generation time of the pixel-wise image generator and for DADA, the search is only conducted on a small proxy subdataset.

\begin{figure}[ht]
    \centering
    \includegraphics[width=0.95\linewidth]{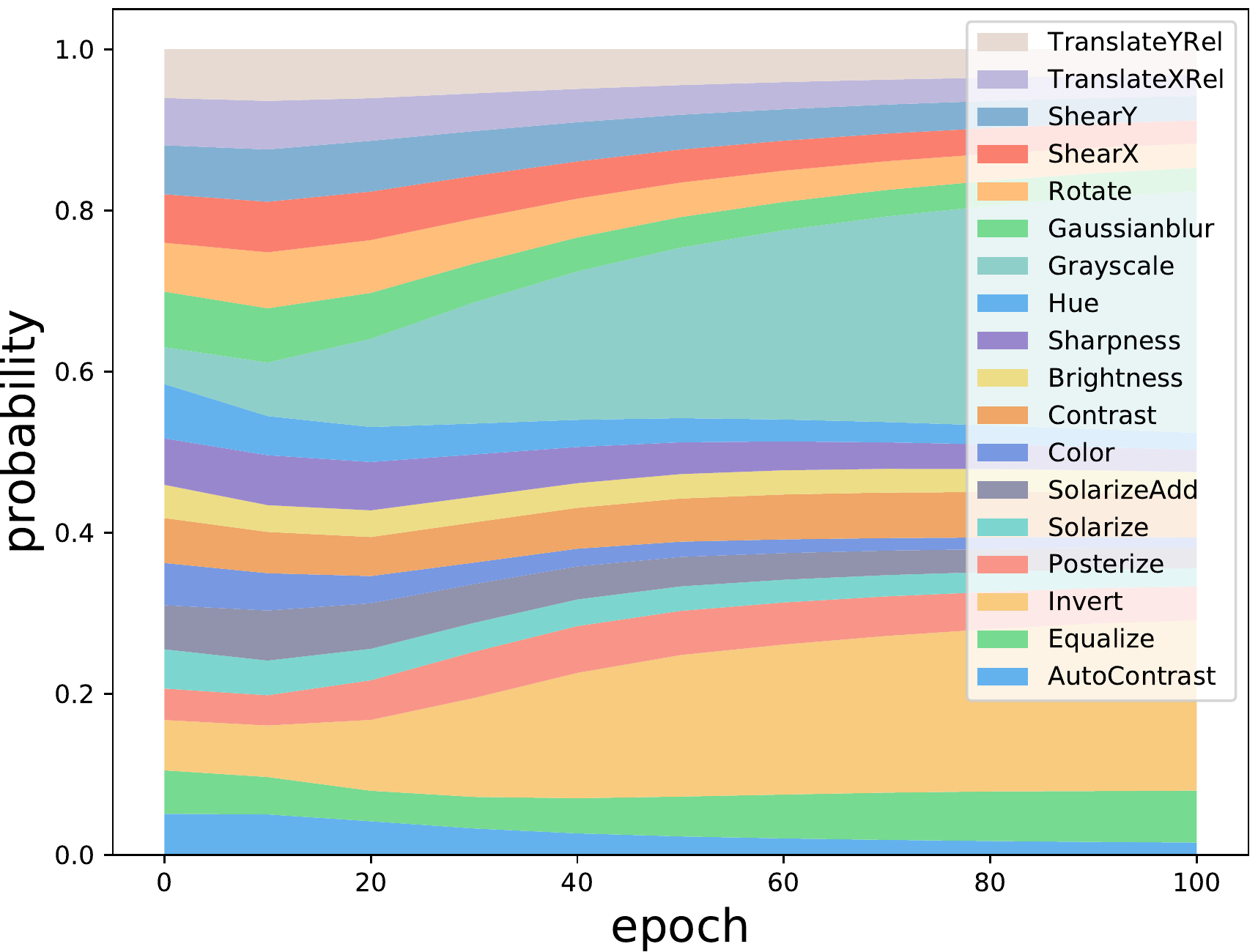}
    \caption{Operations' sampling probability on the training process.}
    \label{fig:op_weight}
\end{figure}

\textbf{Qualitative Studies of AutoView Searching.}
To demonstrate the effectiveness of our AutoView searching, we visualize the different operations' sampling probability searched in \cref{fig:op_weight}.  In particular, we put geometric transformation (e.g.,ShearY, TranslateX) back into our search space in this experiment. We can see that the probabilities of geometric transformation operations are getting smaller while GrayScale, the augmentations commonly used in self-supervised methods, is getting larger, which proves the correctness of our AutoView. 

\begin{figure*}[ht]
\centering
\includegraphics[width=1.0\linewidth]{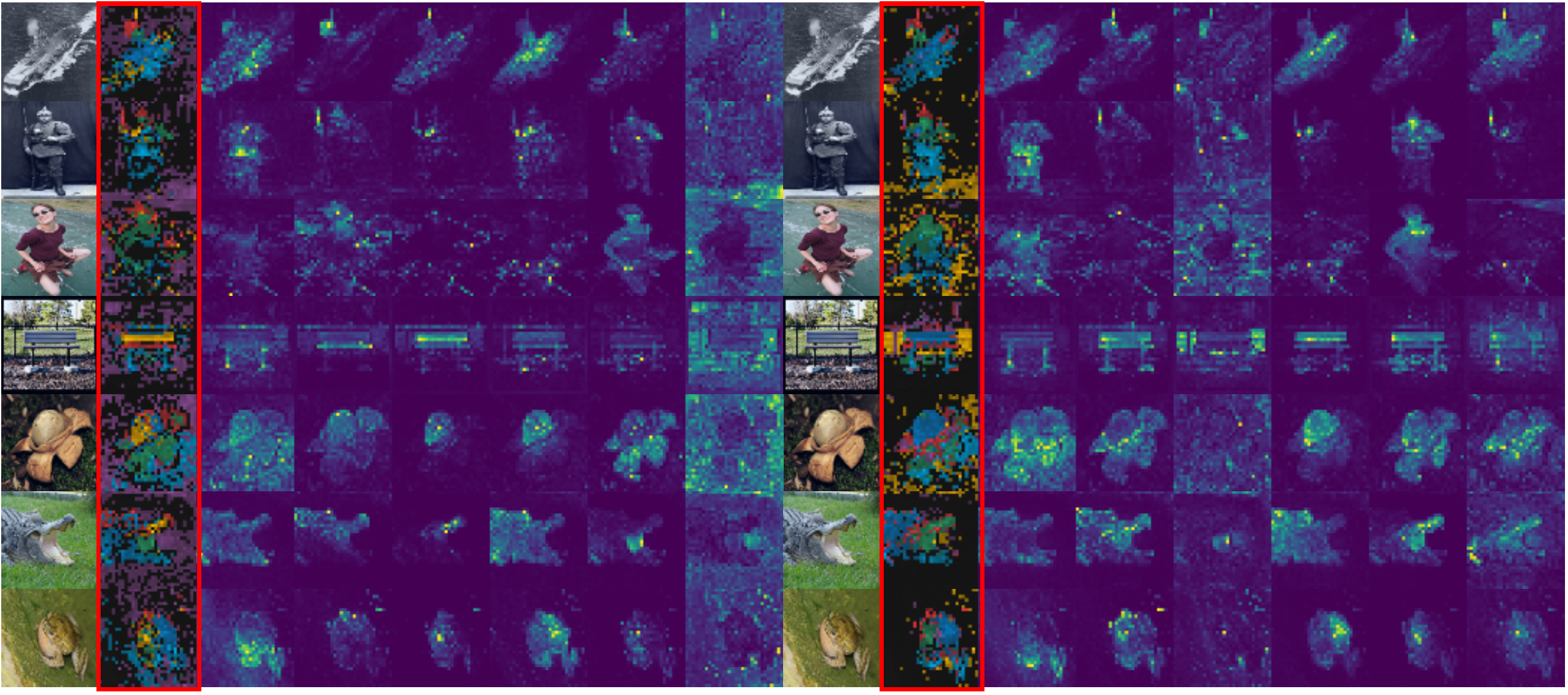}
\caption{Visualization for self-attention map from multiple heads of a ViT-S/16 trained for 100 epochs with AutoView, without (left) and with (right) our self-regularized loss.
}
\label{fig:attnmap2}
\end{figure*}



\textbf{Visualization Results of Attention Maps}
In \cref{fig:attnmap1}, we indicate that our AutoView shows visually stronger ability to capture the information of objects, especially in situations with complex backgrounds, compared with DINO Manual. 
For example, in the first column and the last column, the attention maps learned by AutoView are the complete information of objects (e.g., cats and cabinets) and eliminates the interference of background as much as possible than DINO Manual.

As \cref{fig:attnmap2} shows, without self-regularized loss, the features learned by the model are chaotic. By comparison, the model trained with our self-regularized loss can learn the target object information well.

\section{Conclusion} \label{sec:conclusion}
In this work, we present AutoView, a self-regularized adversarial AutoAugment method, to learn views for self-supervised vision transformers. Extensive experiments have shown that AutoView achieves remarkable improvements over RandAug baseline (up to 10.2\% on $k$-NN classification), and consistently outperforms view policies designed by human experts on various tasks. Proved by ablation studies, AutoView can outperform RandAug with grid search by 2.4\% linear accuracy on same search space. And our curated search space has been proved to improve the performance of RandAug baseline significantly by 6.5\% linear accuracy. Ablation analyses have also shown that our proposed self-regularized loss term successfully addressed the problem of information collapse. Additionally, experiments on the efficient self-supervised learning scheme by progressive training also demonstrate the effectiveness and generalization of AutoView.


\ifCLASSOPTIONcaptionsoff
  \newpage
\fi



\bibliographystyle{IEEEtran}
\bibliography{IEEEabrv,egbib}

\end{document}